\documentclass{article}

\usepackage{microtype}
\usepackage{graphicx}
\usepackage{booktabs} 

\usepackage{color}
\usepackage{amssymb}
\usepackage{nameref}
\usepackage{subcaption}
\usepackage{graphicx}

\usepackage{hyperref}
\usepackage{amsmath}

\usepackage[accepted]{icml2018}

\icmltitlerunning{Learning under selective labels in the presence of expert consistency}

\begin{document}

\twocolumn[
\icmltitle{Learning under selective labels in the presence of expert consistency}



\icmlsetsymbol{equal}{*}

\begin{icmlauthorlist}
\icmlauthor{Maria De-Arteaga}{ml,aut,he}
\icmlauthor{Artur Dubrawski}{aut}
\icmlauthor{Alexandra Chouldechova}{he,stat}
\end{icmlauthorlist}

\icmlaffiliation{ml}{Machine Learning Department}
\icmlaffiliation{aut}{Auton Lab}
\icmlaffiliation{he}{Heinz College}
\icmlaffiliation{stat}{Department of Statistics}

\icmlcorrespondingauthor{Maria De-Arteaga}{mdeartea@andrew.cmu.edu}

\icmlkeywords{Machine Learning, ICML}

\vskip 0.3in
]



\printAffiliationsAndNotice{}  

\begin{abstract}
We explore the problem of learning under selective labels in the context of algorithm-assisted decision making. Selective labels is a pervasive selection bias problem that arises when historical decision making blinds us to the true outcome for certain instances. Examples of this are common in many applications, ranging from predicting recidivism using pre-trial release data to diagnosing patients. In this paper we discuss why selective labels often cannot be effectively tackled by standard methods for adjusting for sample selection bias, even if there are no unobservables. We propose a data augmentation approach that can be used to either leverage expert consistency to mitigate the partial blindness that results from selective labels, or to empirically validate whether learning under such framework may lead to unreliable models prone to systemic discrimination.
 
\end{abstract}

\section{Introduction}

In many domains, humans are routinely tasked with making predictions to inform decisions their job requires them to make. Examples of such are judges who predict the likelihood of recidivism when determining bail, and doctors who predict the likelihood of neurological recovery of comatose patients when deciding whether to extend life support. Increasingly, machine learning is being used to aid humans in those predictions. Research has extensively shown that machine learning and actuarial models are better at making predictions than humans~\cite{meehl1954clinical,dawes1989clinical,grove2000clinical}, but the available data frequently presents an overlooked challenge: Human decisions often determine whether the true outcome (label) is observed. For example, when a judge does not grant bail, we are blind to the counterfactual of what would have happened if the individual had been released. 

This problem has been recently formalized as the selective labels problem~\cite{kleinberg2017human,lakkaraju2017selective}. Under this setting, if machine learning algorithms are trained using the observed outcomes, the resulting models are not answering the question ``given an individual $x_i$, is situation $Y$ likely to occur?'', but rather, ``given an individual $x_i$ for whom a human predicts that situation $Y$ is likely to occur, is situation $Y$ indeed likely to occur?''. Thus, rather than estimating the quantity $P(Y=1|X)$, the learning algorithms can be viewed as estimating $P(Y|X,D=1)$, where $Y$ denotes the true label, $X$ the covariates available for prediction, and $D$ the human decision. 

In this paper we argue that under selective labels the intimate relationship between the source of the selection bias---humans predicting the target label---and the target label itself makes one of the core assumptions of existing methods to correct for selection bias no longer reasonable, as we cannot assume that every instance has a non-negligible probability of being labeled. For example, in the criminal justice setting there may be conditions under which an offender is ineligible for bail. In New Hampshire, for example, defendants may be ineligible for pre-trial release if they are arrested for murder or for an alleged violation of a domestic violence protective order, among others~\cite{bailcommissionerhandbook_2016}. 

\begin{figure*}[h!]
\begin{center}
\includegraphics[width=0.77\linewidth]{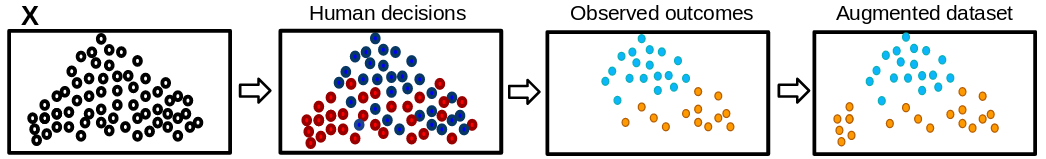}
\end{center}
\caption{Diagram illustrating proposed data augmentation.}
\label{fig:diagram}
\end{figure*}

We propose a way of leveraging consistency; whenever we are blind to the true label for a portion of the population, it is because humans consistently predict that these cases belong to one class. As a result, we can make use of experts' knowledge to partially overcome the blindness. This allows us to learn from humans in the cases where they are very confident---and where it would be impossible for the machine to learn without introducing other assumptions---and learn from observational data in all other cases. Figure \ref{fig:diagram} illustrates the approach. The proposed data augmentation can be used to learn what humans already seem to know whenever consistency is believed to stem from expertise, and it can be used as an evaluation framework whenever it is not clear whether human consistency is indicative of correctness.

\section{Related Literature}
\label{sec:lit}

\citet{kleinberg2017human,lakkaraju2017selective} propose a way of better evaluating algorithmic predictions in the presence of both selective labels and unobservables. Our work adds to the discussion they initiated by identifying an additional limitation of applying existing methodology to address selective labels. Additionally, while their methodology considers the problem of model \textit{validation} and leverages the \textit{heterogeneity} of human decisions by focusing on those cases where humans disagree, our focus is instead on the \textit{homogeneity} of human decisions, its risks and opportunities for model \textit{training}.  

The fairness-related risks of learning from censored data are explored in~\citet{kallus2018residual}. The selective labels problem is a special case of \textit{sample selection bias}, which concerns learning in a setting where training and test data are drawn from different distributions~\citep{zadrozny2004learning,huang2007correcting}. Statistics and quantitative methods literature on missing data has also addressed this problem~\citep{little2014statistical, seaman2013review}. In addition to assuming conditional ignorability, which fails in the presence of unobservables, a common assumption to the different approaches that have been presented to tackle sample selection bias is that every individual has a non-zero probability of being part of the training sample, i.e., $P(d_i=1|x_i) > 0$ $\forall i=1,..,n$, where $d_i$ refers to $x_i$ being selected for the sample. This so-called positivity assumption is rarely reasonable to make under selective labels, as individuals who are very easy for humans to label in the class that makes us blind to the true output will never appear in our labeled data. If the algorithm never observes the true label for a subpopulation, it cannot learn anything about that group without making further assumptions. If trained only on observed outcomes, what it learns for that subset will be by extrapolation, which will not necessarily be correct. 



Our work shares similarities with the literature on learning to defer~\cite{cortes2016learning, madras2017predict}, which also combines human and algorithmic decision making. However, existing techniques in this realm, which rely on the algorithm's ability to self-assess its performance and confidence, are not directly applicable, because under selective labels we cannot assess the accuracy of humans or machines for the unlabeled cases we are concerned about. We do note, however, that a framework for learning to defer using the selection bias as a criteria would be a plausible alternative to the proposed methodology.  

\section{Data Augmentation}
\label{sec:meth}

A primary risk of learning under selective labels is that those cases that are very easy for humans to predict, and whose accurate prediction blinds us to the true outcome, will never be seen by the algorithm during training.

In this section we propose a simple data augmentation scheme for using expert certainty to overcome some of the blindness that results from selective labels. The core assumption of our methodology is that whenever humans are consistent in a region of the feature space, their prediction is likely to be correct. It is important to note that we are not assuming humans will display overall consistency, which is not generally a reasonable assumption when predictions involve human behavior \citep{shanteau1992competence, shanteau2015task}. Rather, we are saying that when humans display consistency for a subset of cases, however big or small that portion may be, we assume their consistency is indicative of correctness, an assumption that is often reasonable when decision-makers are knowledgeable domain experts. In the pre-trial release context, for instance, this subset may consist of bail ineligible defendants, who all judges would be observed to detain.

Assume our dataset is composed of triplets $(\mathbf{x_i},d_i,y_i) $, where $\mathbf{x}_i\in \mathbb{R}^n$ refers to a vector of feature values of instance $i$, $d_i \in \{0,1\}$ to the human decision, and $y_i\in \{0,1\}$ to the target variable. The human decision $d_i=1$ allows us to observe the true outcome, while $d_i=0$ does not.  Finally, we assume that the human's decision $d_i$ is based on their prediction of $y_i$. When learning a model to predict $y$, the data available for training is that shown in Equation \ref{eq:D_o}. 
\begin{eqnarray}\label{eq:D_o}
S_{o}=\{(\mathbf{x_i},y_i) : d_i=1\}\\
\label{eq:D_A}
S_{A}=S_o \cup \{(\mathbf{x_i},d_i) : P(d_i=1|x_i)<\epsilon\}
\end{eqnarray}

By learning a predictive model of the human decisions, we can obtain estimates for $P(d_i=1|x_i)$, which corresponds to the probability that the true outcome will be observed. Unlike the task of predicting the true outcome, predicting the human decisions does not present a selective labels challenge. Therefore, in general circumstances this model can be evaluated through standard techniques. Assuming this model has good performance, $P(d_i=1|\mathbf{x_i})<\epsilon$ indicates that humans will consistently label this instance as $d_i=0$, hence the probability for the true label of $x_i$ to be observed in our dataset is negligible. For such instances, we propose to augment our dataset by accepting the label predicted by the humans as the true outcome, resulting in the dataset shown in Equation \ref{eq:D_A}.


The resulting dataset no longer suffers from complete blindness in any portion of the feature space, and the remaining sample selection bias can be corrected through standard techniques. For instance, one can apply inverse probability weighting \citep{little2014statistical}, weighting instance $i$ by $w_i \equiv 1 / P(d_i = 1 | \mathbf{x_i})$.  Note that such strategies are generally not valid if the selection is based on unobservables that are predictive of the target label; i.e., they still rely on the assumption that $Y \perp D \mid X$.  


\section{Experiments}
\label{sec:exp}

Allegheny County child maltreatment hotline receives over 15,000 calls a year.  Call workers are tasked with deciding whether a case warrants further investigation, giving rise to a selective labels setting. The County is working on developing risk assessment models to assist call workers in their decisions.  In this section we analyze the potential risk of learning under selective labels in this setting and explore the effectiveness of the proposed data augmentation methodology, both in real and semi-synthetic data. 

The dataset consists of all 83,311 referrals associated to a total of 47,305 children received between 2010 and 2014. For each referral, there are over 800 variables available, including information regarding demographics, behavioral health, and past interactions with county prison and public welfare for all adults and children associated to a referral. For each case we know whether the call was screened-in by the caseworker, and if screened in, we observe whether an out-of-home placement occurred in the 730 days following the call. In the experiments below, we consider the first referral associated to each child, and train random forest models with a $75\%$-$25\%$ train-test partition. The calibration of predicted probabilities is tested, as seen in Figure~\ref{fig:cal}.

\subsection{Semi-synthetic labels}

We simulate a case in which humans have high recall but not necessarily high precision, which is consistent with what would be desirable in this and other settings, where we may expect humans to investigate whenever there is a slim suspicion. For example, doctors prescribe blood tests knowing that a large portion will test negative for the conditions of interest. Similarly, without resource constraints, we might expect workers to screen in calls whenever there is a slight suspicion of child abuse. To achieve this, we assign semi-synthetic labels $d^s$ and $y^s$, shown in Equation~\ref{eq:semisynt_d}.

\begin{equation}
\begin{array}{rcl}
d^s_i & = & \left\{ \begin{array}{rcl}
d_i & \mbox{for} & P(d_i=1|\mathbf{x_i}) > 0.9\\
0 & \mbox{for} & P(d_i=1|\mathbf{x_i}) \leq 0.9 \\
\end{array}\right.  \\
y^s_i & = & \left\{ \begin{array}{rcl}
y_i & \mbox{for} & P(d_i=1|\mathbf{x_i}) > 0.9\\
0 & \mbox{for} & P(d_i=1|\mathbf{x_i}) \leq 0.9 \\
\end{array}\right.
\end{array}
\label{eq:semisynt_d}
\end{equation}

\subsection{Assessing the impact of selective labels}

\begin{figure}[ht!]
\begin{subfigure}[c]{0.22\textwidth}
    \includegraphics[width=\linewidth]{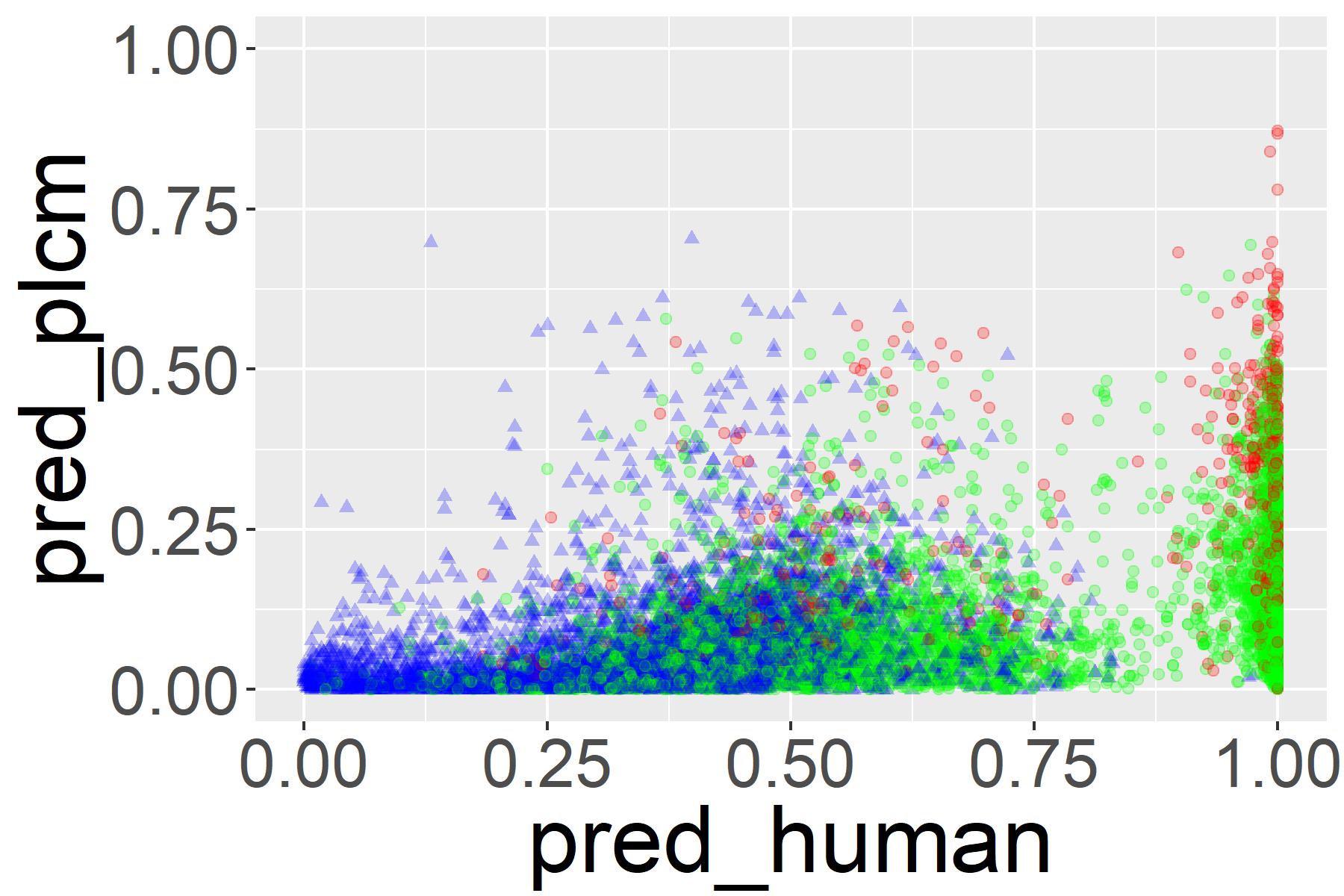}
    \caption{Real-world, observed.}
    \label{scatter:real_obs}
  \end{subfigure}
  \hfill
  \begin{subfigure}[c]{0.22\textwidth}
   \includegraphics[width=\linewidth]{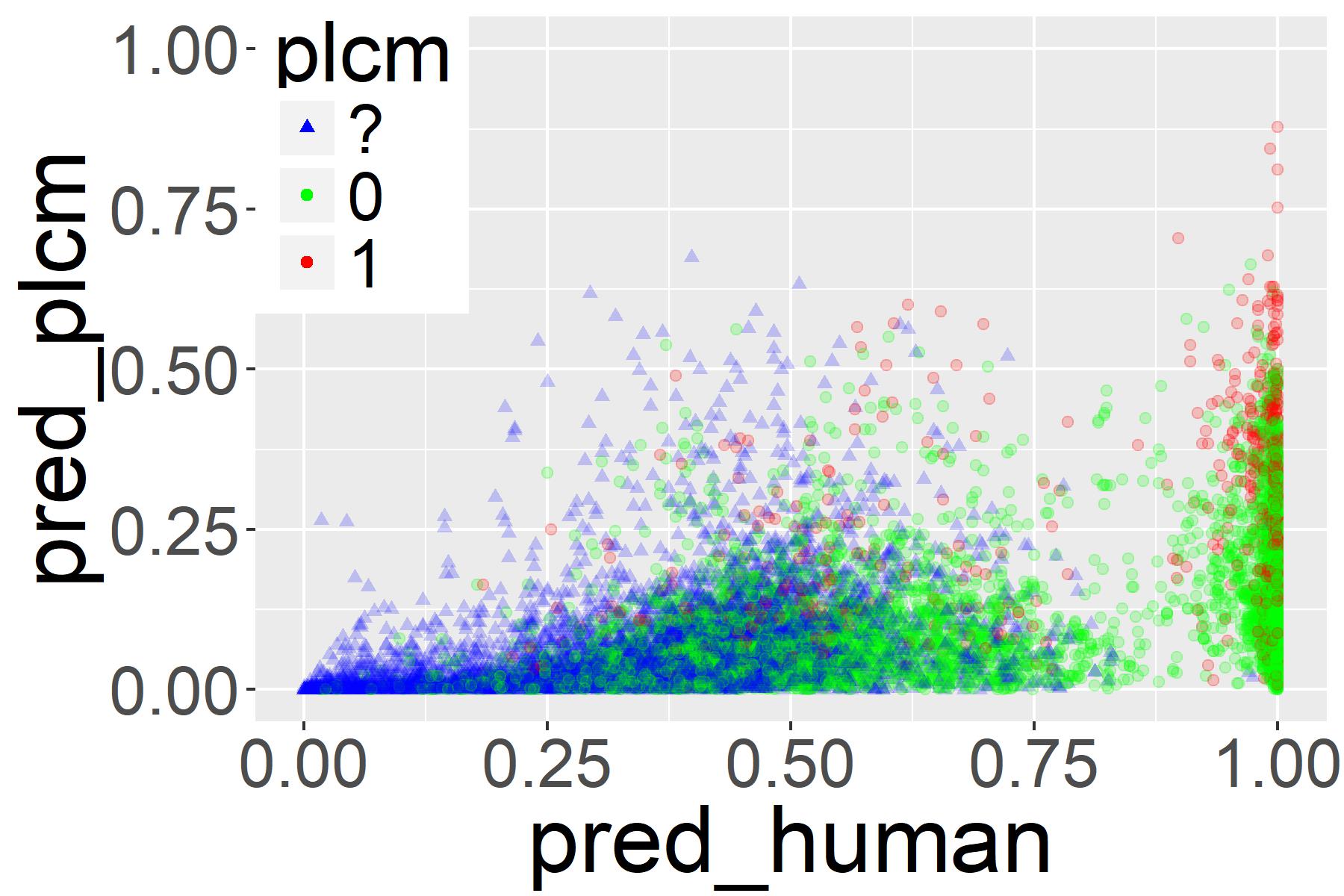}
    \caption{Real-world, augmented.}
    \label{scatter:real_augm}
  \end{subfigure}
  \begin{subfigure}[c]{0.22\textwidth}
    \includegraphics[width=\linewidth]{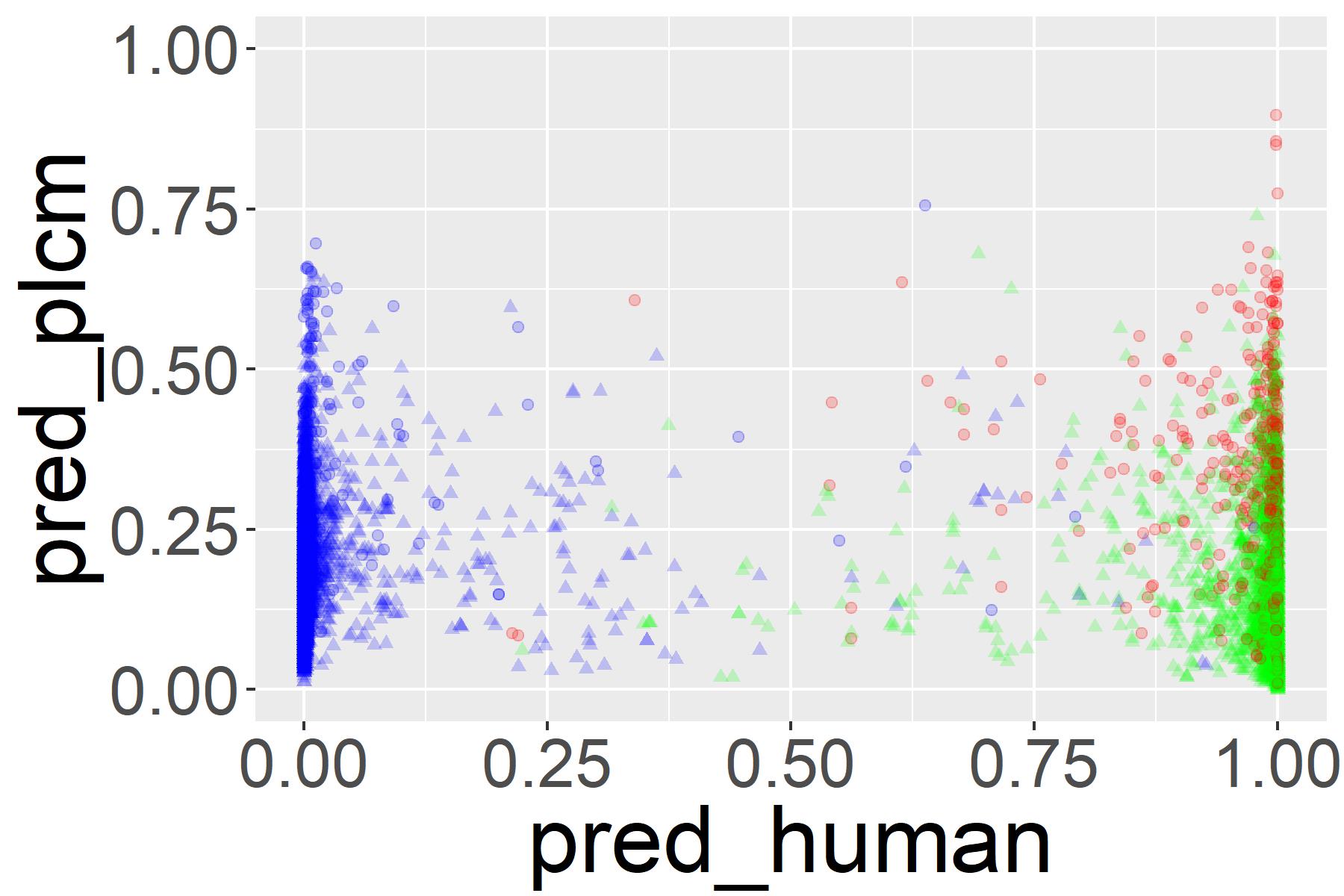}
    \caption{Semisynthetic, observed.}
    \label{scatter:synt_obs}
  \end{subfigure}
  \hfill
  \begin{subfigure}[c]{0.22\textwidth}
   \includegraphics[width=\linewidth]{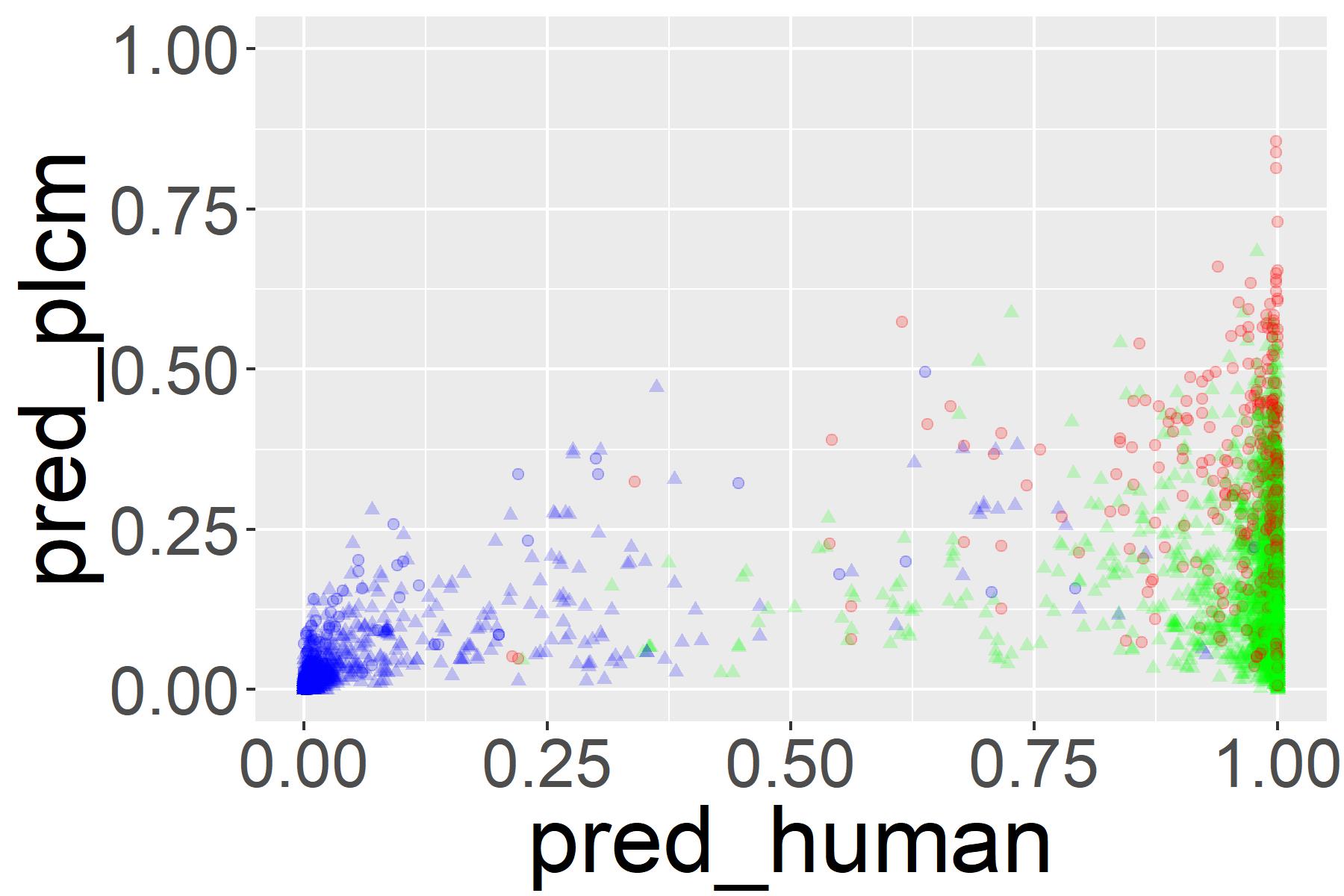}
    \caption{Semisynthetic, augmented.}
    \label{scatter:synt_augm}
  \end{subfigure}
  \caption{Predictions of human decision, $d$, and predictions of placement, $y$, for the test sets. Captions specify training set. Colors indicate if placement occurred (when observed) and `?' denotes the case was screened-out hence the label was not observed.}
  \label{fig:scatter}
\end{figure}

We can empirically assess some of the potential impacts of learning under selective labels by comparing the model trained to predict the human decisions and the model trained to predict placement using screened-in cases. Figures \ref{scatter:real_obs} and \ref{scatter:synt_obs} compare the risk scores assigned by the two, each displayed on one axis. The information conveyed in these plots is helping in assessing the potential impact of learning under selective labels. If the placement model's predictions do not match the human models' predictions in the cases that are confidently screened out---the left-most area of the scatterplot---it means one of the two is wrong. For cases where decision makers are knowledgeable experts, it is plausible that this is due to the placement model not generalizing well to the entire dataset. 

In the real-world data, it can be observed that the placement model assigns a relatively low probability to the cases that are confidently screened-out, indicating a high degree of consistency between the two models. In the semi-synthetic data, however, we can observe that what is learned from the observed data does not generalize to those cases that are confidently (and correctly) screened out.  

\begin{figure}
  \begin{subfigure}[c]{0.22\textwidth}
   \includegraphics[width=\linewidth]{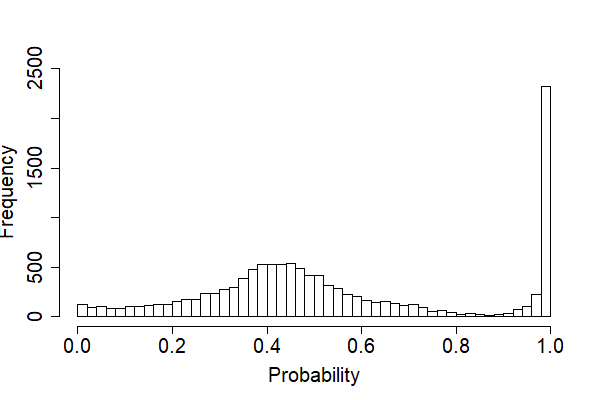}
    \caption{Predicted probabilities.}
    \label{fig:hist}
  \end{subfigure}
    \begin{subfigure}[c]{0.22\textwidth}
   \includegraphics[width=\linewidth]{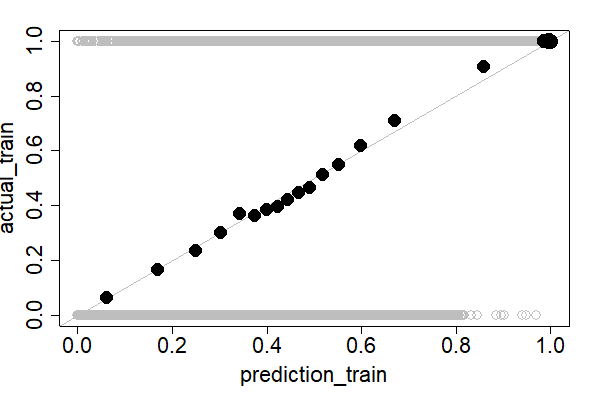}
    \caption{Calibration verification.}
    \label{fig:cal}
  \end{subfigure}
  \caption{Random forest predictions to estimate $P(D=1\mid X)$.}
\end{figure}

\subsection{Data augmentation}

\begin{figure}[ht!]
\begin{subfigure}[c]{0.22\textwidth}
    \includegraphics[width=0.95\textwidth]{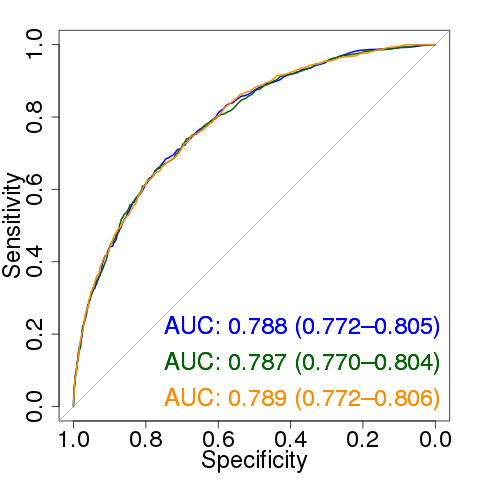}
    \caption{Real-world, observed.}
    \label{fig:eval_obs_real}
  \end{subfigure}
  \hfill
  \begin{subfigure}[c]{0.22\textwidth}
    \includegraphics[width=0.95\textwidth]{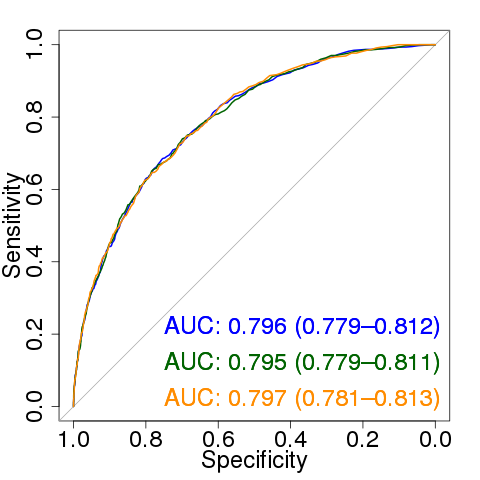}
    \caption{Real-world, augmented.}
    \label{fig:eval_augm_real}
  \end{subfigure}
    \begin{subfigure}[c]{0.22\textwidth}
    \includegraphics[width=0.95\textwidth]{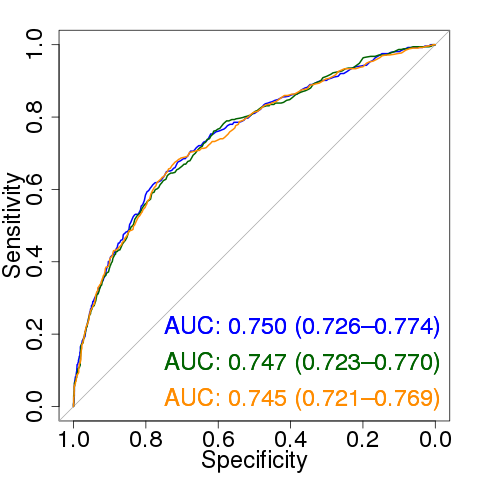}
    \caption{Semisynthetic, observed.}
    \label{fig:eval_obs_synt}
  \end{subfigure}
  \hfill
  \begin{subfigure}[c]{0.22\textwidth}
    \includegraphics[width=0.95\textwidth]{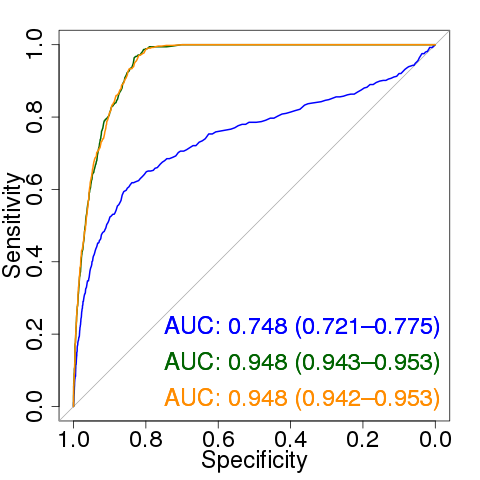}
    \caption{Semisynthetic, augmented.}
    \label{fig:eval_augm_synt}
  \end{subfigure}
  \caption{ROC curve of model trained only on observed data (blue), on augmented data (green) and on augmented data with inverse probability weights (yellow). Captions indicate evaluation test set.}
  \label{fig:roc_comp}
\end{figure}

We apply the proposed data augmentation technique with $\epsilon=0.05$, training the model on the augmented dataset. In Figures \ref{scatter:real_augm} and \ref{scatter:synt_augm} we can observe that individuals who are predicted to be confidently screened-out are assigned a lower risk by the algorithms trained on augmented data, with this effect being much more prominent in the semi-synthetic data. In Figure \ref{fig:roc_comp} we evaluate three models: (1) trained only on observed outcomes, (2) trained on augmented data, and (3) trained on augmented data and using inverse probability weights to correct for remaining selection bias. The evaluation is done in the portion of the test set for which we observe true outputs as well as in the augmented test set. 
The fact that there is no difference in overall performance in terms of ROC for the real-world data is likely explained by the fact that while a significant number of screen-ins are predicted with high-confidence, only a small amount of screen-outs are predicted with high confidence, as observed in Figure \ref{fig:hist}. This results in very little difference between the observed and augmented sets. In the case of semi-synthetic data, the data augmentation successfully incorporates human knowledge, as observed in Figure \ref{fig:eval_augm_synt}. Moreover, Figure \ref{fig:eval_augm_synt} reveals that the model that is trained only on observed outcomes performs very poorly at low sensitivity (false negative rate), and a comparison with Figure and~\ref{fig:eval_obs_synt} shows that this can only be detected when evaluating on the augmented data. 

\section{Discussion}
\label{sec:disc}

One of our main empirical findings is that while data augmentation may have little to no impact on the performance of the model on the labeled data, it can have a large effect on model generalization to the selectively unlabeled region of the data. Beyond the gains that this represents in terms of the model's reliability, increased agreement between the model predictions and confident human decisions may help to build trust in the model in decision support contexts. Human assessors are likely to form a negative impression of a tool if it frequently issues a contradictory assessment in cases where the human is confident. This may make the human assessors less likely to trust the model in general, and thus could undermine its utility even in cases where the model is far more confident and accurate than the human.

Another way in which learning from the confidence of experts can help to build trust is through attenuating the effects of omitted payoff bias. In making decisions, call workers are assessing the referral for a broader set of potential risks and harms than simply out of home placement. Confident human decisions may be thought of as a proxy for some of these other outcomes. The model trained on the augmented $Y$ is predicting a hybrid target. A direction for future work consists on exploring whether such hybrid prediction indeed increases trust in the system.   

A note must be made regarding unobservables---features that are available to the humans, but which are not captured in the data and thus are unavailable to the machine. This problem would neither be solved nor exacerbated by the proposed data augmentation method. Cases for which unobservables lead humans to make differing decisions will never be a part of the augmented portion. The model may still learn incorrect associations from the labeled data, but will not learn any additional incorrect associations from the augmented data. 

In future work we plan to explore the risks of learning under selective labels when human consistency is not indicative of correctness, but rather a consequence of shared misconceptions or biases. Additionally, we intend to study the theoretical guarantees of the proposed methodology, while also exploring ways of accounting for unobservables.

\section*{Acknowledgements}
We thank Dr. Kyle Miller for his fruitful suggestions. This research was supported in part by the Metro21 Institute at Carnegie Mellon University, and DARPA under awards FA8750-1720130 and FA8750-1420244.  

\bibliography{bib}
\bibliographystyle{icml2018}
\end{document}